\documentclass{neus2026}

\title[RT-NeuS]{RT-NeuS: Towards Real-Time Neuro-Symbolic Video Understanding via Adaptive Temporal Verification}
\usepackage{times}

\usepackage{microtype, wrapfig}
\usepackage{comment}
\usepackage{graphicx}
\usepackage{subcaption}
\usepackage{booktabs} 

\usepackage{hyperref}
\usepackage{algorithmic}
\usepackage{algorithm}
\usepackage{url}
\usepackage{multirow, nicefrac}
\usepackage{pifont}
\usepackage{xcolor}
\usepackage{colortbl}
\usepackage{makecell}  
\usepackage{wrapfig} 
\usepackage{enumitem}
\usepackage{booktabs}
\usepackage{todonotes}

\author{%
 \Name{Shawn Liang$^1$} \Email{shawn.liang@case.edu}\\
 \vspace{-16pt}
 \AND
 \Name{Sahil Shah$^2$} \Email{ss96869@utexas.edu}\\
  \vspace{-16pt}
 \AND
 \Name{Chengwei Zhou$^1$} \Email{chengwei.zhou@case.edu}\\
  \vspace{-16pt}
 \AND
 \Name{S P Sharan$^2$} \Email{spsharan@utexas.edu}\\
  \vspace{-16pt}
 \AND
 \Name{Harsh Goel$^2$} \Email{harshg99@utexas.edu}\\
  \vspace{-16pt}
 \AND
 \Name{Arnab Sanyal$^2$} \Email{sanyal@utexas.edu}\\
  \vspace{-16pt}
 \AND
 \Name{Sandeep Chinchali$^2$} \Email{sandeepc@utexas.edu}\\
  \vspace{-16pt}
 \AND
 \Name{Gourav Datta$^1$} \Email{gourav.datta@case.edu} \\
  \vspace{4pt}
  $^1$\textit{Case Western Reserve University}, \ \ $^2$ \textit{The University of Texas at Austin}
  \vspace{-16pt}
}

\begin{document}

\maketitle
\begin{abstract}
Long-form video question answering (LVQA) requires answering natural-language queries about videos spanning minutes to hours, demanding temporal reasoning across thousands of frames. Standard vision-language models (VLMs) struggle with this task: their fixed frame budgets force aggressive downsampling that misses the temporal structure that complex queries depend on. Neuro-symbolic approaches address this by decomposing queries into atomic propositions, translating them into temporal logic specifications, and applying formal model checking to retrieve segments satisfying the specification. This yields up to 10\% higher accuracy on temporally complex benchmarks, with interpretability and formal guarantees. However, constructing the video automaton requires grounding every proposition at every frame window via VLM calls, resulting in up to $130\times$ slower inference than standard VLM prompting. We present RT-NeuS, a framework that preserves the accuracy and formal guarantees of temporal-logic-guided LVQA while closing this latency gap. RT-NeuS introduces coarse-to-fine adaptive sampling to identify the small set of query-relevant, visually distinct frames, and batched proposition detection with KV-cache reuse to evaluate all propositions per window in a single forward pass. We derive latency upper bounds as a function of video length, proposition count, and sampling density. Experiments on LongVideoBench, Video-MME, and MLVU reduce inference latency by up to $13\times$ on a single NVIDIA H200 GPU, while matching or exceeding prior neuro-symbolic accuracy.
\end{abstract}

\section{Introduction}
\begin{figure*}[t]
  \centering
  \includegraphics[width=1\linewidth]{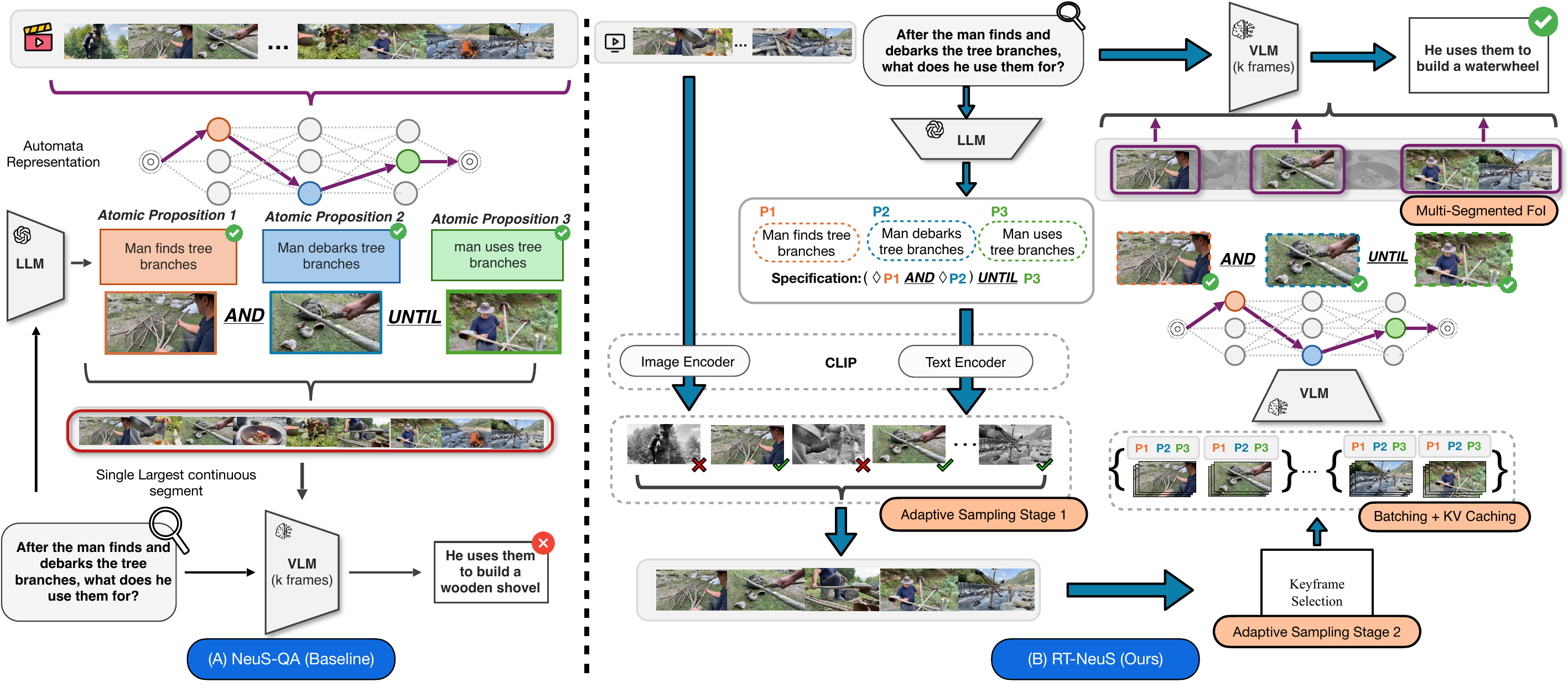} 
\caption{\footnotesize (a) \textbf{NeuS-QA (Baseline)} densely grounds atomic propositions, constructs temporal logic specifications, and retrieves a single logic-consistent segment through formal model checking, but incurs high sequential latency. (b) \textbf{RT-NeuS (Ours)} introduces \textit{Coarse-to-Fine Adaptive Sampling} and \textit{Batched Proposition Detection with KV-Cache Reuse} to selectively ground propositions over semantically distinct keyframes. \textit{Multi-Segment FoI Retrieval} extracts multiple disjoint logic-consistent segments for final VQA. Together, RT-NeuS achieves a $13\times$ end-to-end speedup while improving accuracy on temporally complex queries and preserving formal guarantees.}

\label{fig:overview}
\end{figure*}

\noindent Real-world applications such as surveillance, dash cam analysis, and autonomous vehicles increasingly rely on hour-long videos, requiring systems that can respond in seconds rather than minutes. However, achieving long-form video question answering (LVQA) poses a trade-off between reasoning quality and inference speed, making it a primary bottleneck in the field. To illustrate, consider a one-hour long wilderness vlog and the following question, ``\textit{After the man finds and debarks the tree branches, what does he use them for?}'' (Figure~\ref{fig:overview}). To reason about the answer to this query, the system must perform semantic grounding (identify the man, the branches, the actions), temporal reasoning (locate events across long durations), and compositional inference (assemble events into a chain matching the query structure)~\citep{shah2025neusqa,choi2024neuro, shah2025challenge}.

Recent vision-language models (VLMs) have demonstrated impressive accuracy on video question answering (VQA) tasks with short video clips \citep{bai2025qwen, opengvlab2024internvl, achiam2023gpt4}. Traditional VLMs follow the approach of uniformly sampling frames to obtain visual context for question-answering. With this approach, the model is forced to aggressively downsample to avoid exceeding token limits on longer videos, leaving large chunks of the video undersampled, thereby risking missing subtle event transitions and temporal cues that are crucial for answering temporal queries~\citep{park2026lvnet,shah2025neusqa,choi2024neuro}. Neuro-symbolic approaches such as NeuS-QA~\citep{shah2025neusqa} address this by translating natural language queries into temporal logic (TL) specifications and applying formal model checking to retrieve segments that satisfy the query \citep{shah2025neusqa,choi2024neuro,ye2025tstar}. On LongVideoBench, TL-based models outperform traditional VLMs by up to 10\% on temporally-complex queries~\citep{shah2025neusqa}. In addition, they offer interpretability and formal guarantees that heuristic-based retrieval methods lack~\citep{park2026lvnet, wang2024videoagent, wang2025videotree, ye2025tstar}. However, these suffer from an end-to-end inference speed that is almost $130\times$ slower than that of traditional VLMs. Figure~\ref{fig:overview} illustrates this trade-off: Uniform sampling and heuristic retrieval have lower latencies but lack interpretability and formal guarantees; while NeuS-QA offers formal guarantees at the cost of significantly increased latency due to the video automata construction.

In this work, we present RT-NeuS: an efficient neuro-symbolic method that retains high accuracy and formal guarantees while offering dramatically lower inference latency. RT-NeuS utilizes the following three components:
\begin{enumerate}[label=(\alph*)]
    \item \textbf{Coarse-to-Fine Adaptive Sampling}: a two-stage sampling that first identifies semantically significant segments and then removes visual-duplicates within these segments, reducing grounding redundancy.
    \item \textbf{Batched Proposition Detection with KV-Cache Reuse}: a single batched VLM query that uses a single visual KV-cache to share visual representations among multiple propositions, therefore evaluating all propositions in each window just once per window. 
    \item \textbf{Multi-Segment Frames-of-Interest (FoI) Retrieval}: Extracting several disjoint segments satisfying the logical formula, maximizing the evidential frames in the visual contexts fed into the final VQA. 
\end{enumerate}
Our contribution is twofold: Firstly, we show that our RT-NeuS framework is able to achieve comparable accuracies and retain formal guarantees while offering substantially lower inference latencies. We evaluate RT-NeuS on LVQA benchmark datasets LongVideoBench~\citep{wu2024longvideobench}, Video-MME~\citep{fu2024videomme}, and MLVU~\citep{zhou2025mlvubenchmarkingmultitasklong} using four state-of-the-art VLM backbones. Second, we offer the first systematic analysis of latency in neuro-symbolic video understanding and propose principled optimizations that enable a practical path toward real-time deployment.

\section{Related Work}
\noindent\textbf{Video Question Answering}: Recent advances in VLMs~\citep{bai2025qwen, opengvlab2024internvl, achiam2023gpt4, li2024llava, li2024aria} have significantly improved zero-shot capabilities and generalizability in VQA tasks with short videos. Despite these advances, such models rely on fixed frame sampling, limiting their ability to reason over long-form videos with complex temporal structure. A separate line of work tackles long-form video through specialized architectures~\citep{weng2024longvlm, li2024videomamba} or system-level inference optimizations~\citep{zhang2025flashvstream, schneider2026quickvideo}, improving efficiency within a monolithic VLM pipeline. Retrieval-augmented models reduce irrelevant context using heuristics or text-based summaries~\citep{park2026lvnet, wang2024videoagent, ye2025tstar, wang2025videotree, song2024moviechat, islam2025bimba, choudhury2023zeroshot}, but sacrifice visual fidelity and lack formal reasoning guarantees.
\begin{wrapfigure}{r}{0.4\textwidth}
  \centering
\includegraphics[width=0.4\textwidth]{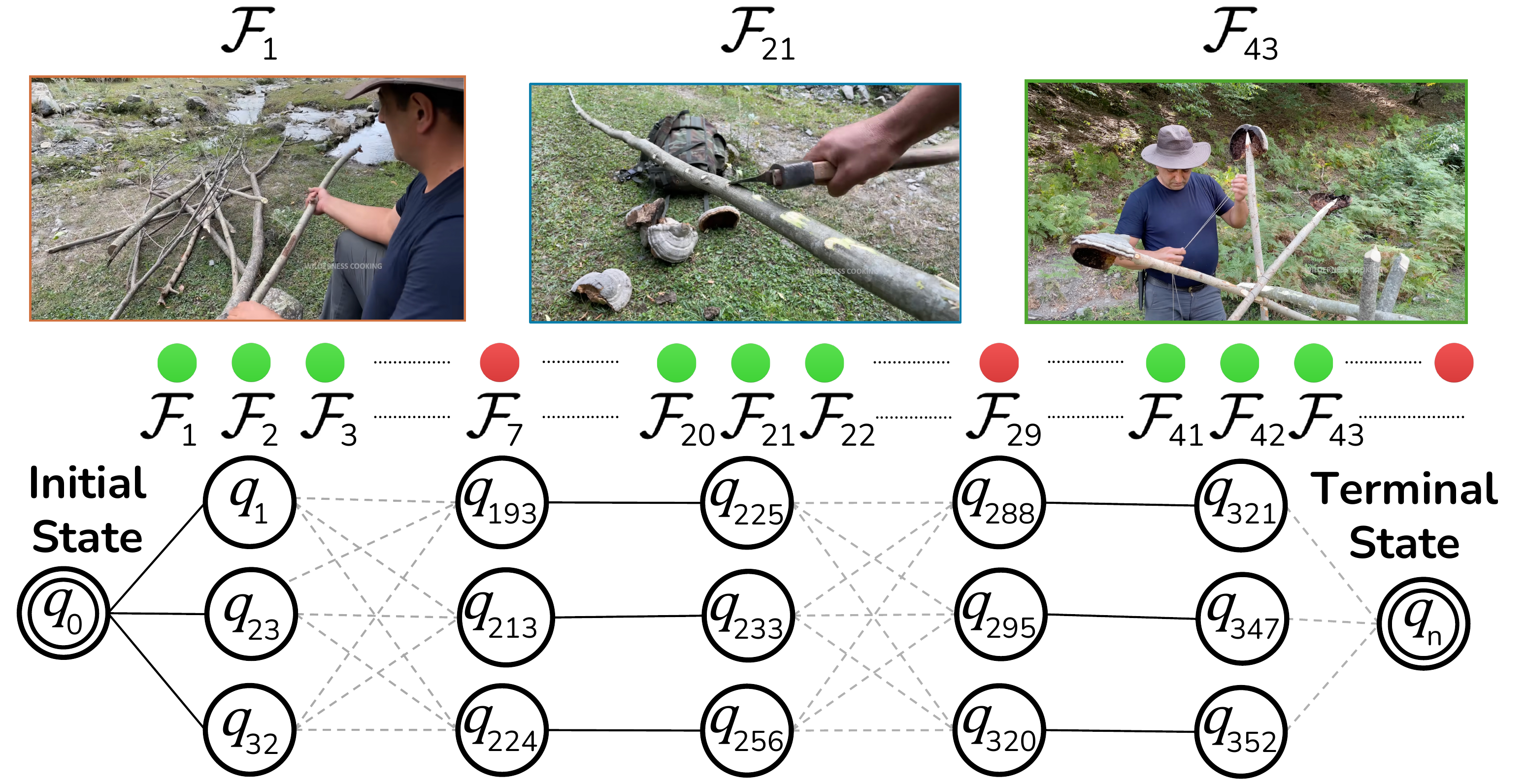}
\caption{\footnotesize [Adapted from our prior work \citet{shah2025neusqa}] Video automaton for the running example. \textbf{Top}: representative frames for key events. \textbf{Bottom}: finite-state automaton from $\varphi$. Green states satisfy temporal constraints; red states denote violations. Incremental verification ensures that only $\rho \models \varphi$ runs are retained, and reaching the accepting state yields a segment satisfying the query.
}
  \label{fig:automaton_example}
  \vspace{-5mm}
\end{wrapfigure}

\noindent\textbf{Neuro-Symbolic Video Understanding}: Prior work has applied symbolic representation to autonomous vehicles~\citep{zheng2025neurostrata, mehdipour2023formal, shoukry2017linear}, robotics~\citep{puranic2021learning, hasanbeig2019reinforcement}, and visual recognition~\citep{manginas2024nesya, hashemi2023neurosymbolic}. In video understanding, previous works utilize symbolic techniques for long-form video~\citep{tapaswi2016movieqa, huang2020movienet, xiong2019graph, wu2021towards}. However, these methods typically rely on latent embeddings, lacking interpretability. To address this issue, recent work integrates temporal logic with neural perception for video understanding. ~\cite{choi2024neuro} introduced the NSVS-TL, which decouples perception from temporal reasoning through video automata. ~\cite{yang2023specification} developed specification-driven video search combining foundation models and formal verification. ~\cite{sharan2025neusv} proposed text-to-video evaluation using symbolic temporal logic, demonstrating 5$\times$ higher correlation with human judgments over prior metrics. ~\cite{choi2025post} showed how neuro-symbolic feedback improves video generation quality by 40\%. Most recently, NeuS-QA~\citep{shah2025neusqa} achieved state-of-the-art LVQA accuracy through temporal logic-guided segment retrieval, improving over 10\% on temporally complex queries in LongVideoBench~\citep{wu2024longvideobench}. Despite these advances, none address the latency bottleneck of neuro-symbolic pipelines, which is the focus of this work.

\section{Preliminaries and Problem Formulation}
\subsection{Neuro-Symbolic Video Understanding}
We adopt the temporal-logic-guided LVQA framework of NeuS-QA~\citep{shah2025neusqa, choi2024neuro}.  For a natural-language query $q$, a set of atomic propositions $\mathcal{P} = \{p_1, \ldots, p_n\}$ is derived (\textit{e.g.}, for our running example: \textit{man finds tree branches}, \textit{man debarks tree branches}, \textit{man uses tree branches}). A temporal logic specification $\varphi$ is built from $\mathcal{P}$ using temporal operators such as $\Diamond$ (eventually) and $\mathcal{U}$ (until)~\citep{clarke1999model,huth2004logic}; for our example: $\varphi = (\Diamond p_1 \land \Diamond p_2) \ \mathcal{U} \ p_3$. The video $V$ of length $T$ is modeled as a discrete-time Markov chain~\citep{norris1998markov,kemeny1960finite}: $\mathcal{A}_V = (Q, q_0, \delta, \lambda)$, where each state $q_i \in Q$ corresponds to a frame window $\mathcal{F}_i$ of $\kappa$ consecutive frames, $\delta: Q \times Q \to [0,1]$ is the transition probability, and $\lambda: Q \to 2^{\mathcal{P}}$ labels states with detected propositions. Given $\mathcal{A}_V$ and $\varphi$, the Storm model checker~\citep{hensel2020storm} computes the satisfaction probability $P_{\text{sat}} = \mathbb{P}[\mathcal{A}_V \models \varphi]$ and returns a satisfying path $\pi = q_{t_1} q_{t_2} \cdots q_{t_k}$ that identifies a video segment passed to the final VLM for answer generation. Figure~\ref{fig:automaton_example} illustrates the construction.

\subsection{Latency Decomposition and Bottleneck Analysis}
\label{sec:latency_analysis}
We decompose the end-to-end cost of the pipeline into its functional stages.
\begin{definition}[NeuS Pipeline Latency]
The end-to-end latency $\mathcal{L}_{\text{NeuS}}$ of a neuro-symbolic video understanding pipeline decomposes additively into four stages: $\mathcal{L}_{\text{NeuS}} = \mathcal{L}_{\text{LQ2TL}} + \mathcal{L}_{\text{auto}} + \mathcal{L}_{\text{MC}} + \mathcal{L}_{\text{VQA}}$, where $\mathcal{L}_{\text{LQ2TL}}$, $\mathcal{L}_{\text{auto}}$, $\mathcal{L}_{\text{MC}}$, and $\mathcal{L}_{\text{VQA}}$ represent the latencies for query-to-TL translation, automaton construction, model checking, and final VQA answering, respectively.
\end{definition} 
Three stages are approximately constant in $T$, collectively denoted as the fixed overhead $\mathcal{L}_{\text{fixed}} = \mathcal{L}_{\text{LQ2TL}} + \mathcal{L}_{\text{MC}} + \mathcal{L}_{\text{VQA}}$. Only $\mathcal{L}_{\text{auto}}$ scales linearly with $T$, since automaton construction requires VLM grounding over every frame window.

\begin{definition}[Sequential Automaton Construction Latency]
Given $\mathcal{A}_V$ with frame windows of size $\kappa$, the sequential automaton construction latency is: $\mathcal{L}_{\text{auto}}^{\text{seq}} = N \cdot |\mathcal{P}| \cdot \mathcal{L}_{\text{prop}}$, where $N=\left\lceil \frac{T}{\kappa} \right\rceil$ is the number of windows and $\mathcal{L}_{\text{prop}}$ the latency of a single proposition-window evaluation.
\label{def:automaton}
\end{definition}
\noindent As $T$ grows, $\mathcal{L}_{\text{auto}}^{\text{seq}}$ shifts to dominate the pipeline.
It exceeds the fixed overhead whenever $T > T_{\text{crit}}$:
\begin{align}
    \mathcal{L}_{\text{auto}}^{\text{seq}} &> \mathcal{L}_{\text{fixed}} \\
    \left\lceil \frac{T_{crit}}{\kappa} \right\rceil \cdot |\mathcal{P}| \cdot \mathcal{L}_{\text{prop}} &= \mathcal{L}_{\text{fixed}}\\
    T_{\text{crit}} &= \frac{\kappa \cdot \mathcal{L}_{\text{fixed}}}{|\mathcal{P}| \cdot \mathcal{L}_{\text{prop}}}
\end{align}

\section{Proposed Method}
We propose RT-NeuS, a latency-efficient neuro-symbolic framework that preserves the formal verification of temporal-logic-guided LVQA while removing the main latency bottleneck of prior pipelines. RT-NeuS exploits two properties of long-form video: semantic sparsity (most frames are irrelevant to a given query) and visual redundancy (neighboring relevant frames are visually similar), together with the fact that proposition grounding pairs the same visual context with every atomic proposition. The full pipeline is given in Algorithm~\ref{alg:rtneus} and illustrated in Figure~\ref{fig:overview}; the following subsections describe each component in detail.

\subsection{Batched Proposition Detection with KV-Cache Reuse} NeuS-QA evaluates each (window, proposition) pair as a separate VLM forward pass, re-encoding the same visual input $|\mathcal{P}|$ times per window. RT-NeuS reduces the redundancy at two levels.

\begin{definition}[Proposition Detection with Shared Visual Prefix] Let $\mathcal{F}_t$ denote the frame window and $\mathcal{P} = \{p_1, \ldots, p_{|\mathcal{P}|}\}$ denote the atomic propositions. RT-NeuS computes proposition confidences for $\mathcal{F}_t$ in two steps. First, the vision encoder runs once on $\mathcal{F}_t$ to produce visual tokens, which are prefilled to build a visual KV-cache $\mathbf{K}_v$. Then, each proposition $p_i$ is evaluated by prefilling only its textual tokens against $\mathbf{K}_v$, producing a confidence score $z_{t,i} \in [0,1]$. The $|\mathcal{P}|$ text prefills are batched into a single forward pass over the shared visual prefix. 
\label{def:batched_detection}
\end{definition} 
The per-window cost reduces from $|\mathcal{P}|$ full forward passes to one visual prefill plus one batched text-only prefill, with the visual KV stored once rather than duplicated across propositions.

\subsection{Coarse-to-Fine Adaptive Sampling}
Batched proposition improves per-window throughput, but the number of windows still grows linearly with the length of the video. To break this linear scaling, we exploit the \textbf{Semantic sparsity} and \textbf{Visual Redundancy} described above with a two-stage adaptive sampling. For a given query, only a small fraction of frames support any proposition in $\mathcal{P}$; in an hour-long video, relevant content may span only a few minutes. Within these relevant segments, consecutive frames are visually similar, so grounding each proposition independently yields duplicate labels and redundant VLM calls.
We utilize a two-stage filter using CLIP embeddings~\citep{radford2021clip} as a lightweight proxy, restricting VLM inference to only the relevant and distinctive frames. All embeddings are $\ell_2$-normalized for mathematical simplicity.

\vspace{4pt}
\noindent\textbf{Stage 1: Semantic Relevance Filtering}
$\text{CLIP}_V$ denotes the visual encoder and $\text{CLIP}_T$ denotes the text encoder.

\begin{definition}[Relevancy Score]
\label{def:relevancy}
Let $\mathbf{z}_{f_t} = \frac{\text{CLIP}_V(f_t)}{\|\text{CLIP}_V(f_t)\|_2}$ and $\mathbf{z}_{p_i} = \frac{\text{CLIP}_T(p_i)}{\|\text{CLIP}_T(p_i)\|_2}$ be the $\ell_2$-normalized visual and textual embeddings. The relevance of frame $f_t$ to the proposition set $\mathcal{P}$ is the maximum cosine similarity over all propositions:
$s_t = \max_{p_j \in \mathcal{P}} \mathbf{z}_{f_t}^\top \mathbf{z}_{p_j}.$
\end{definition}
Frames with $s_t > \tau_s$ are retained, each extended into a temporal window of radius $w$, and overlapping windows are merged into the candidate set $\mathcal{F}_{cand} \subseteq V$. Stage 1 is a recall-first filter: it excludes frames clearly unrelated to any proposition before VLM inference. This allows us to set $\tau_s$ conservatively towards recall: false positives may incur additional Stage 2 and VLM compute but are caught by downstream filtering, while false negatives are unrecoverable since the discarded frames never reach grounding. We aggregate with $\max$ rather than $\text{mean}$ because propositions describe disjoint events: averaging would dilute the matching proposition's score with noise from non-matching ones.

\vspace{4pt}
\noindent\textbf{Stage 2: Visual Redundancy Elimination}: Stage 2 eliminates visually similar frames within $\mathcal{F}_{cand}$, retaining only visually distinct keyframes.
\begin{definition}[Visual Redundancy Score]
For any two frames $f_a, f_{b} \in \mathcal{F}_{cand}$, the redundancy score $r$ is the cosine similarity of their normalized CLIP visual embeddings: $r(f_a, f_{b}) = \mathbf{z}_{f_a}^\top \mathbf{z}_{f_{b}}$, both $z_{f_a}$ and $z_{f_{b}}$ are image embeddings as defined in Definition ~\ref{def:relevancy}.
\end{definition}
\begin{definition}[Sequential Keyframe Selection]
Let $\mathcal{F}_{cand} = \{f_1, f_2, \dots, f_M\}$ be the temporally ordered candidate frames. We construct the keyframe set $\mathcal{K}$ iteratively: Initialize $\mathcal{K} = \{f_1\}$ and set the current comparison base $f_{base} = f_1$; for each subsequent frame $f_t \in \mathcal{F}_{cand}$, we compute $r(f_{base}, f_t)$; if $r(f_{base}, f_t) < \tau_r$, add $f_t$ to $\mathcal{K}$ and update comparison base $f_{base} \leftarrow f_t$, otherwise discard $f_t$.
\end{definition}
A run of visually similar frames collapses to a single representative, the anchor advances only when a sufficiently different frame appears. The threshold $\tau_r$ governs precision on redundancy: values close to 1 admit only near-identical frames as redundant, while lower values propagate more aggressively. VLM grounding is performed over the neighborhood $\mathcal{F}_{\text{detect}} = \bigcup_{k \in \mathcal{K}} \{f_t : |t - k| \leq \delta\}, \ \delta = 2.$ Frames in $\mathcal{F}_{cand} \setminus \mathcal{F}_{\text{detect}}$ inherit labels from their preceding keyframe, having been discarded only because their redundancy score exceeded $\tau_r$. Frames in $\mathcal{V} \setminus \mathcal{F}_{cand}$ receive zero labels, reflecting that no proposition is likely to hold. CLIP visual embeddings capture scene-level changes (cuts, camera motion, object presence) but miss fine-grained pose or interaction changes, which the neighborhood expansion recovers.


\vspace{6pt}
\noindent\textbf{Multi-Segment FoI Retrieval}: A verified segment can span a long interval while containing little evidence. When a query involves $A \ \mathcal{U} \, B$, the verifier may return an interval spanning both events plus the filler between them, diluting the evidence ratio under a fixed frame budget. Instead, RT-NeuS returns the intersection of the verified span with the positive proposition regions, concatenated in temporal order. These segments preserve the verified temporal sequence while removing the intermediate filler, giving the VLM an evidence-dense visual input of the same verified path. Formally, let $|E|$ be the number of evidence frames in the returned context, and $|\mathcal{V}_{\text{ret}}|$ the total frame count of the returned context. The probability that the VLM samples at least one evidence frame is $P(\text{Hit}) \approx 1 - \left(1 - \frac{|E|}{|\mathcal{V}_{\text{ret}}|}\right)^{N_{\text{VQA}}}$. The benefit grows with the temporal spread: long-horizon queries with large verified spans see the largest reduction in FoI size and the greatest gain in evidence density.

\begin{algorithm}[t]\footnotesize
\caption{RT-NeuS: Towards Real-Time Neuro-Symbolic Video Understanding}
\label{alg:rtneus}
\begin{algorithmic}[1]
\renewcommand{\algorithmicrequire}{\textbf{Input:}}
\renewcommand{\algorithmicensure}{\textbf{Output:}}
\REQUIRE Video $V = \{f_1, \dots, f_T\}$, query $q$, batch size $B$, thresholds $\{\tau_s, \tau_r\}$, window $w$, segment offsets $(w_{\text{pre}}, w_{\text{post}})$
\ENSURE Answer $\hat{c}$

\STATE $\mathcal{P}, \varphi \gets \texttt{LQ2TL}(q)$ \hfill \COMMENT{Translate query to Temporal Logic}
\STATE $\{z_{f_t}\}_{t=1}^T \gets \texttt{CLIP}_V(V)$ \hfill \COMMENT{Batched image encoding}
\STATE $\mathcal{F}_{cand} \gets \{f_t : \max_{p \in \mathcal{P}} \cos(z_t, \text{CLIP}_T(p)) > \tau_s\}$ \hfill \COMMENT{Phase 1: Semantic Filtering}

\STATE $\mathcal{K} \gets \{f_{1}\}, \ f_{base} \gets f_1$ \hfill \COMMENT {Phase 2: Visual redundancy elimination}
\FOR{each $f_t \in \mathcal{F}_{cand}$}
    \IF{$\cos(z_{base}, z_t) < \tau_r$}
        \STATE $\mathcal{K} \gets \mathcal{K} \cup \{f_t\}$, \ $f_{base} \gets f_t$
    \ENDIF
\ENDFOR
\STATE $\mathcal{F}_{\texttt{detect}} \gets \bigcup_{k \in \mathcal{K}} \{f_t : |t-k| \leq \delta\}$ \hfill \COMMENT{Expand keyframes to detection neighborhoods}

\STATE Initialize $Z \in [0,1]^{T \times |\mathcal{P}|}$
\FOR{each window $F_t$ centered at $f_t \in \mathcal{F}_{\text{detect}}$ with size $\kappa$}
    \STATE $Z[t, :] \gets \texttt{VLM}_{\texttt{batch}}(f_t, \mathcal{P})$ \hfill \COMMENT{Batched detection with KV-cache reuse}
\ENDFOR 
\STATE $Z[t, :] \leftarrow Z[t_{\text{prev-keyframe}}, :] for \ t \in \mathcal{F}_{cand} \setminus \mathcal{F}_{\text{detect}}$ \hfill \COMMENT{Label propagation}
\STATE $\mathcal{A}_V \gets \texttt{BuildAutomaton}(Z)$ \hfill \COMMENT{Symbolic grounding}
\STATE $P_{\text{sat}}, \pi^* \gets \texttt{Storm}(\mathcal{A}_V, \varphi)$ \hfill \COMMENT{Formal model checking}
\STATE $S \gets \text{ExtractMultiSegment}(\pi^*, Z, w_{\text{pre}}, w_{\text{post}})$ \hfill \COMMENT{Multi-segment FoI retrieval}
\STATE $\hat{c} \gets \texttt{VLM}_{\texttt{answer}}(q, S)$ \hfill \COMMENT{Final multimodal reasoning}
\STATE \Return $\hat{c}$
\end{algorithmic}
\end{algorithm}

\section{Theoretical Analysis}
We derive an upper bound on the end-to-end latency of RT-NeuS and use it to characterize when latency efficiency is achievable. We let $\mathcal{L}_{\text{VLM}}$ denote the per-window latency of batched proposition detection with KV-cache reuse: one visual prefill to produce the cached KV prefix, plus one batched text-only prefill over all $|\mathcal{P}|$ propositions (Definition~\ref{def:batched_detection}). We let $\mathcal{L}_{\text{prop}}$ denote the latency of one sequential proposition-window evaluation in the baseline. On batching-friendly hardware, $\mathcal{L}_{\text{VLM}} \approx \mathcal{L}_{\text{prop}}$ since the visual encoder dominates per-call cost; in practice $\mathcal{L}_{\text{VLM}} < \mathcal{L}_{\text{prop}}$ because KV-cache reuse eliminates redundant visual encoding.

\begin{theorem}[RT-NeuS Latency Bound]
\label{thm:latency}
Let $\alpha = \frac{|\mathcal{F}_{cand}|}{T}$ be the semantic retention rate, $\rho = \frac{|\mathcal{K}|}{|\mathcal{F}_{cand}|}$ the keyframe retention rate, and $\mathcal{L}_{\text{CLIP}}$ the per-frame CLIP encoding latency. The end-to-end latency of RT-NeuS is bounded by
\begin{equation}
\mathcal{L}_{\text{RT-NeuS}} \leq \mathcal{L}_{\text{LQ2TL}} + T \cdot \mathcal{L}_{\text{CLIP}} + \left\lceil \frac{\alpha \rho T}{\kappa} \right\rceil \cdot \mathcal{L}_{\text{VLM}} + \mathcal{L}_{\text{MC}} + \mathcal{L}_{\text{VQA}},
\end{equation}
where T, $\kappa$, and the per-stage latencies $\mathcal{L}_{\text{LQ2TL}}$, $\mathcal{L}_{\text{MC}}$, $\mathcal{L}_{\text{VQA}}$ are as defined in section~\ref{sec:latency_analysis}.
\end{theorem}

\begin{corollary}[Condition for Latency Efficiency]
\label{cor:realtime}
let $L_{fixed}$ denote the fixed overhead, and $N_{\text{win}} = \lceil \frac{T}{\kappa} \rceil$ the total window count. Then RT-NeuS achieves ($\mathcal{L_{\text{RT-NeuS}}} \leq \mathcal{L}_{\text{max}}$) whenever 
$$\alpha \rho \leq \frac{\mathcal{L}_{\text{max}} - \mathcal{L}_{\text{fixed}} - T \cdot \mathcal{L}_{\text{CLIP}}}{ N_{\text{win}} \cdot \mathcal{L}_{\text{VLM}}}$$
\end{corollary}
The condition isolates $\alpha\rho$, the fraction of total windows processed by the VLM, as the controllable variable. Fixed costs and CLIP encoding set an irreducible floor; the cascade reduces $\alpha\rho$ to fit within the remaining budget.
\begin{proposition}[Speedup over Baseline]
\label{prop:speedup}
Let $\mathcal{L}_{\text{auto}}^{\text{seq}}$ denote the baseline NeuS-QA automaton-construction latency (Definition ~\ref{def:automaton}).
Assuming $\mathcal{L}_{\text{VLM}} \approx \mathcal{L}_{\text{prop}}$, the speedup on the automaton-construction stage is $$\frac{\mathcal{L}_{\text{auto}}^{\text{seq}}}{\mathcal{L}_{\text{auto}}^{\text{RT-NeuS}}} \approx \frac{|\mathcal{P}|}{\alpha \rho} \cdot \frac{1}{1 + \frac{T \cdot \mathcal{L}_{\text{CLIP}}}{\alpha\rho \cdot N_{\text{win}} \cdot \mathcal{L}_{\text{VLM}}}}.$$
\end{proposition}
\noindent Empirically on LongVideoBench, $\alpha\rho \approx 0.194$ on average, ranging from $0.407$ on 15-second videos to $0.181$ on hour-long videos (Appendix~\ref{appendix:retention}). With mean $|\mathcal{P}| \approx 4$, the leading factor $\frac{|\mathcal{P}|}{(\alpha\rho)}$ yields speedup from $\sim 10\times$ on short clips to $\sim 22\times$ on hour-long content. After applying the CLIP-overhead correction, the global predicted speedup is $\sim 14.6\times$. The observed end-to-end speedup of $13.19\times$ (Table~\ref{tab:latency}) falls slightly below this prediction; the gap reflects $\mathcal{L}_{\text{fixed}}$ (LQ2TL, model checking, and final VQA), which does not benefit from the framework. See Appendix~\ref{appendix:speedup} for the full derivation.

\section{Experiments}

\noindent\textbf{Implementation.} We use InternVL2-8B~\citep{opengvlab2024internvl} for proposition grounding during automaton construction, GPT-4o for temporal-logic translation (LQ2TL)~\citep{openai2024gpt4ocard}, and Storm for probabilistic model checking~\citep{hensel2020storm,baier2008principles}. Adaptive sampling uses CLIP ViT-B/32 at 1 fps with thresholds $\tau_s = 0.21$ and $\tau_r = 0.9$. CLIP encoding is performed at a video-length-dependent sampling rate: 2 fps for videos under 60 seconds, decreasing to 1 fps for videos over 600 seconds. This maintains adequate context density for proposition detection on short videos while limiting CLIP throughput requirements on long videos. Propositions are embedded by averaging three template-instantiated phrasings before $\ell_2$ normalization. Final VQA samples $N_{\text{VQA}} = 32$ frames uniformly from the returned segments. All other hyperparameters follow \citet{shah2025neusqa} defaults. For LongVideoBench, subtitles are burned into frames following NeuS-QA. All latency results are measured end-to-end on a single NVIDIA H200 GPU.

\noindent\textbf{Evaluation Backbones \& Baselines.}
We pair RT-NeuS with four VLM backbones for final VQA: InternVL2.5-8B, Qwen2.5-VL-7B, VideoLLaMA3-7B, and LLaVA-Video-7B~\citep{chen2024internvl25,bai2025qwen,damonlpsg2025videollama3,zhang2025llavavideo}. We compare against three categories: (i) base VLMs with uniform frame sampling, (ii) NeuS-QA on the same backbones, and (iii) structured-reasoning frameworks T*~\citep{ye2025tstar}, VideoTree~\citep{wang2025videotree}, VideoAgent~\citep{wang2024videoagent}, and LVNet~\citep{park2026lvnet}. Category (iii) is evaluated on LongVideoBench only. All final VQA experiments use the lmms-eval framework~\citep{zhang2024lmmseval}.

\noindent\textbf{Benchmarks.} Our primary benchmark is LongVideoBench~\citep{wu2024longvideobench}, on which we evaluate against all three baseline categories. Following~\citet{shah2025neusqa}, we focus on the reasoning-intensive subsets T3E, E3E, T3O, and O3O — videos up to 60 minutes that require temporal ordering, causal inference, and compositional reasoning. For additional evaluation, we use Video-MME~\citep{fu2024videomme} (Temporal Reasoning subset) and MLVU~\citep{zhou2025mlvubenchmarkingmultitasklong} (Needle QA and Ego Reasoning subsets), evaluated against base VLMs and NeuS-QA only.

\begin{table*}[t]
\centering
\resizebox{\textwidth}{!}{%
\begin{tabular}{@{}llccccc@{}}
\toprule
 & & \multicolumn{5}{c}{\textbf{LongVideoBench}} \\ \cmidrule(lr){3-7}
\textbf{Framework} & \textbf{Model} & \textbf{T3E} & \textbf{E3E} & \textbf{T3O} & \textbf{O3O} & \textbf{Overall} \\ \midrule
\multirow{2}{*}{\textbf{Base VLM}} & InternVL2\_5-8B~\citep{chen2024internvl25} & 58.90 & 71.28 & 50.00 & 54.55 & 59.61 \\
 & VideoLLaMA3-7B~\citep{damonlpsg2025videollama3} & 60.27 & 61.70 & 58.11 & 50.0 & 57.98 \\
 & LLaVA-Video-7B-Qwen2~\citep{zhang2025llavavideo} & 46.58 & 55.32 & 55.41 & 39.39 & 49.84 \\
 & Qwen2.5-VL-7B-Instruct~\citep{bai2025qwen} & 46.58 & 63.83 & 56.78 & 53.03 & 55.70 \\ \midrule
 \multirow{2}{*}{\textbf{Structured Reasoning Frameworks}} & VideoTree (\cite{wang2025videotree}) & 56.38 & 50.00 & 50.00 & 43.84 & 50.47 \\
 & LVNet (\cite{park2026lvnet}) & 41.94 & 54.17 & 35.48 & 46.15 & 45.59 \\
 & VideoAgent (\cite{wang2024videoagent}) & 45.45 & 52.63 & 27.27 & 27.27 & 40.38 \\
 & T* (\cite{ye2025tstar}) & 28.95 & 35.56 & 48.33 & 22.22 & 35.75 \\ \midrule
\multirow{2}{*}{\textbf{NeuS-QA}} & NeuS-QA + InternVL2\_5-8B & 58.91 & 69.15 & 58.11 & 59.09 & 61.89  \\
 & NeuS-QA + VideoLLaMA3-7B & 56.16 & 68.09 & 59.46 & 62.12 & 61.89  \\
 & NeuS-QA + LLaVA-Video-7B-Qwen2 & 54.80 & 58.51 & 55.41 & 45.48 & 54.07  \\
 & NeuS-QA + Qwen2.5-VL-7B-Instruct & 57.53 & 70.21 & 60.81 & 54.55 & 61.56 \\ \midrule
\multirow{2}{*}{\textbf{RT-NeuS (Ours)}} & InternVL2\_5-8B & 64.39 & 76.60 & 60.81 & \textbf{63.64} & \textbf{67.10} \\
 & RT-NeuS + VideoLLaMA3-7B & 64.38 & \textbf{74.47} & 70.27 & 54.55 & 66.76 \\
 & RT-NeuS + LLaVA-Video-7B-Qwen2 & 57.53 & 61.70 & 50.00 & 45.46 & 54.40 \\
 & RT-NeuS + Qwen2.5-VL-7B-Instruct & \textbf{68.09} & 56.16 & \textbf{64.87} & 65.15 & 63.84 \\ \bottomrule
\end{tabular}%
}
\caption{\footnotesize Top-1 Accuracy (\%) on 4 complex-reasoning subsets of LongVideoBench. We evaluate \textbf{RT-NeuS} against the \texttt{NeuS-QA} baseline and Base VLM inference across four state-of-the-art backbones. \textbf{RT-NeuS} consistently achieves superior performance, notably surpassing \texttt{NeuS-QA} by 5.21\% on LongVideoBench.}
\label{tab:main_results}
\end{table*}

\begin{figure*}[t]
  \centering
  \includegraphics[width=0.9\linewidth]{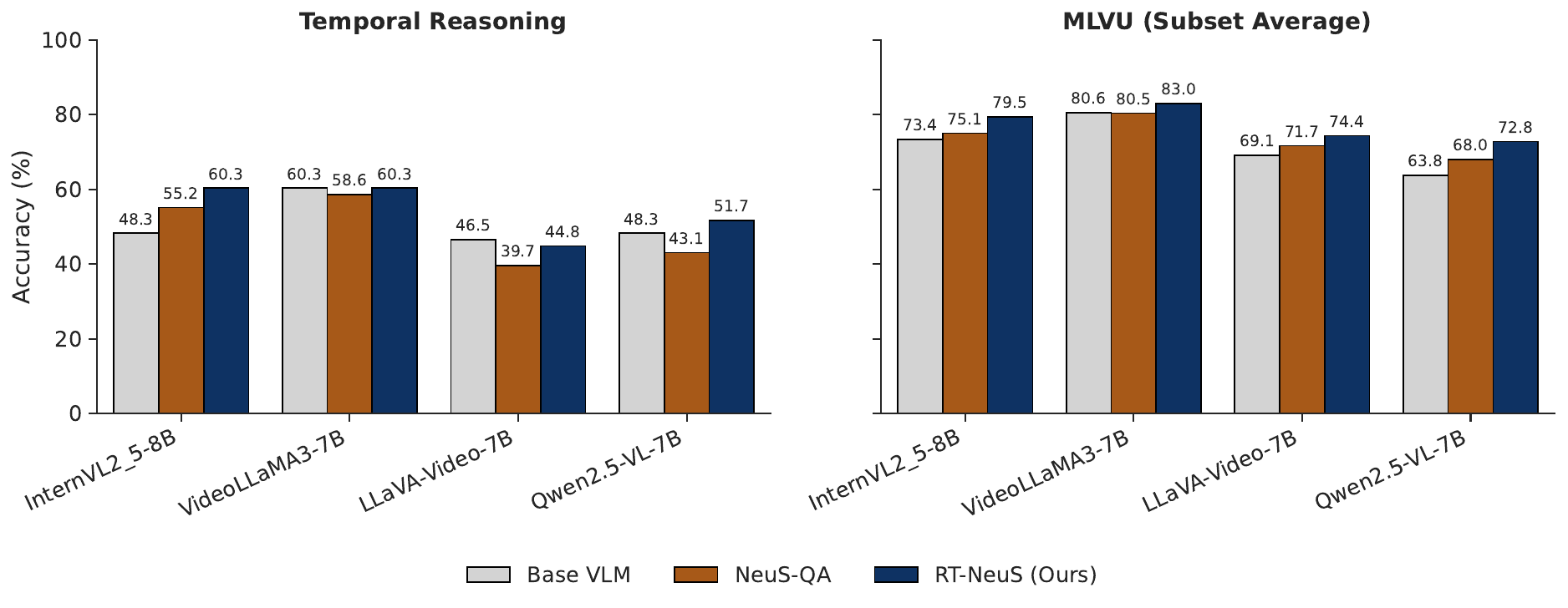}
  \caption{\footnotesize Top-1 Accuracy (\%) on Video-MME, and MLVU. We evaluate \textbf{RT-NeuS} against the \texttt{NeuS-QA} baseline and Base VLM inference for subsets of Video-MME and MLVU. \textbf{RT-NeuS} consistently achieves superior performance, surpassing Base-VLM and NeuS-QA across all question subsets.}
  \label{fig:video-mme-mlvu-accuracy}
\end{figure*}

\noindent\textbf{Accuracy.} RT-NeuS improves over NeuS-QA across all evaluated backbones and benchmarks while substantially exceeding structured-reasoning methods. On LongVideoBench (Table~\ref{tab:main_results}), RT-NeuS with InternVL2.5-8B reaches 67.10\% overall, a 5.21\% accuracy gain over NeuS-QA (61.89\%) and 16.63\% over the best structured-reasoning framework, VideoTree (50.47\%). The superior performance holds across the other three backbones, with RT-NeuS leading NeuS-QA on each. The gains concentrate on temporally complex queries: on Video-MME Temporal Reasoning (Figure~\ref{fig:video-mme-mlvu-accuracy}), RT-NeuS with InternVL2.5-8B reaches 60.3\%, +5.1\% over NeuS-QA baseline. MLVU Needle-QA and Ego-Reasoning show consistent gains across all backbones. The ablation in Table ~\ref{tab:ablation} attributes these gains to Multi-Segment FoI retrieval, which concentrates the final frame sampling on evidence-dense intervals.

\noindent\textbf{Latency.} Table~\ref{tab:latency} compares end-to-end inference latency across video durations. NeuS-QA scales linearly with video length, reaching 935.95s ($\sim 16$ minutes) for hour-long videos. On the same hour-long subset, RT-NeuS completes queries in an average of 64.20s — a 14.58$\times$ speedup. The global average across all durations is 13.19$\times$, with average processed frame count dropping from 824 to 197 and frames-of-interest coverage dropping from 42.81\% to 27.35\% of the video. This gap widens with video length: on hour-long videos, RT-NeuS processes only 9.01\% as FoI versus 24.51\% for NeuS-QA, reflecting the increased semantic sparsity of long-form content.


\begin{table*}[t]
\centering
\resizebox{\textwidth}{!}{%
\label{tab:latency-tab}
\begin{tabular}{l cc cc ccc l}
\toprule
 & \multicolumn{2}{c}{\textbf{Avg. Total Frames Used}} & \multicolumn{2}{c}{\textbf{Avg. FOI \% of Video}} & \multicolumn{3}{c}{\textbf{Avg. Completion Time (s)}} \\ \cmidrule(lr){2-3} \cmidrule(lr){4-5} \cmidrule(lr){6-8}
\textbf{Video Length ($T$)} & \textbf{NeuS-QA} & \textbf{RT-NeuS}  & \textbf{NeuS-QA} & \textbf{RT-NeuS} & \textbf{NeuS-QA} & \textbf{RT-NeuS} & \textbf{Speedup} \\ \midrule
15 (Short) & 44 & 59 & 87.80\% & 77.87\% & 14.4 & 8.09 & 1.78$\times$ \\
60 (Medium) & 71 & 110 & 77.90\% & 62.72\% & 31.39 & 9.93 & 3.16$\times$ \\
600 (Medium-Long) & 422 & 158 & 37.90\% & 19.83\% & 320.27 & 30.45 & 10.52$\times$ \\
3600 (Long) & 1427 & 281 & 24.51\% & 9.01\% & 935.95 & 64.20 & 14.58$\times$ \\ \midrule
\textbf{Global Average} & \textbf{824} & \textbf{197} & \textbf{42.81\%} & \textbf{27.35\%} & \textbf{549.83} & \textbf{41.67} & \textbf{13.19$\times$} \\ \bottomrule
\end{tabular}%
}
\caption{Latency Comparison between standard VLM vs. NeuS-QA vs. RT-NeuS assuming standard VLM uses 32 frames uniform sampling.}
\label{tab:latency}
\end{table*}

\begin{table}[t]
\centering
\resizebox{\columnwidth}{!}{%
\begin{tabular}{l ccc c c c c c }
\toprule
\textbf{Method} & \textbf{Batching} & \textbf{Adaptive} & \textbf{Multi-Seg} & \textbf{Avg. Windows} & \textbf{Avg. VLM Calls} & \textbf{Accuracy} & \textbf{Latency (s)} \\
\midrule
Baseline (NeuS-QA) & - & - & - & 268 & 1206 & 61.89 & 549.83 \\
+ Batched Proposition Detection + KV cache & \checkmark & - & - & 268 & 268 & 61.19 & 151.56 \\
+ Adaptive Sampling & \checkmark & \checkmark & - & \textbf{56} & 56 & 60.18 & \textbf{43.53} \\
\textbf{RT-NeuS (Ours)} & \checkmark & \checkmark & \checkmark & \textbf{56} & 56 & \textbf{67.10} & 41.67 \\
\bottomrule
\end{tabular}%
}
\caption{\textbf{Ablation Study of RT-NeuS Components.} We report the global average VLM window count and latency alongside QA accuracy. Each row adds one component cumulatively.}
\label{tab:ablation}
\end{table}

\label{sec:albation}
\noindent\textbf{Ablation Studies}: Table~\ref{tab:ablation} analyzes each component of RT-NeuS. The baseline exhibits high latency ($549.83$s) due to sequential VLM calls. Adding \textbf{Batched Proposition Detection with KV Caching} reduces latency to $151.56$s ($3.63\times$ speedup) with minimal accuracy change ($61.19\%$). Incorporating \textbf{Adaptive Sampling} further lowers latency to $43.53$s ($3.48\times$ additional speedup) by reducing processed windows from 268 to 56, with a slight accuracy drop ($61.19\% \to 60.18\%$) due to aggressive pruning. Finally, \textbf{Multi-Segment FoI Retrieval} improves accuracy to \textbf{67.10\%} ($+6.92\%$) while maintaining low latency. Overall, RT-NeuS achieves a $13.19\times$ latency reduction and a $+5.21\%$ absolute accuracy gain over the baseline.

\section{Conclusions and Discussions}
We introduce RT-NeuS, a neuro-symbolic framework for long-form video question answering that reduces inference latency from $\sim 130\times$ to $\sim 10\times$ over base VLM prompting while preserving the accuracy and interpretability of temporal logic reasoning. The core observation is that long-form video is semantically sparse and visually redundant relative to a given query, and that proposition grounding pairs the same visual context with every atomic proposition. RT-NeuS exploits both, replacing dense sequential VLM grounding with selective sampling and cache-shared batched inference. Performance remains bounded by VLM-based proposition grounding and scales with specification complexity; replacing the VLM with lightweight, task-specific proposition detectors offers the most direct route toward sub-$3\times$ overhead, paving the way toward real-time neuro-symbolic video understanding.

\section{Acknowledgements}
This material is based upon work supported in part by the
Office of Naval Research (ONR) under Grant No. N00014-
22-1-2254. Additionally, this work was supported by the Defense Advanced Research Projects Agency (DARPA) contract DARPA ANSR: RTX CW2231110. Approved for Public Release, Distribution Unlimited.

\bibliography{neus}
\appendix
\section{Speedup Proposition Derivation}
\label{appendix:speedup}
\begin{proposition}[Speedup over Baseline]
Assuming $\mathcal{L}_{\text{VLM}} \approx \mathcal{L}_{\text{prop}}$, the speedup of RT-NeuS over baseline NeuS-QA on the automaton-construction stage is
$$\frac{\mathcal{L}_{\text{auto}}^{\text{seq}}}{\mathcal{L}_{\text{auto}}^{\text{RT-NeuS}}} \approx \frac{|\mathcal{P}|}{\alpha \rho} \cdot \frac{1}{1 + \frac{T \cdot \mathcal{L}_{\text{CLIP}}}{\alpha\rho \cdot N_{\text{win}} \cdot \mathcal{L}_{\text{VLM}}}}.$$
\end{proposition}

\noindent\textbf{Derivation.} The baseline (Definition~\ref{def:automaton}) processes all $N_{\text{win}} = \lceil\frac{T}{\kappa}\rceil$ windows with $|\mathcal{P}|$ sequential proposition-window evaluations per window:
$$\mathcal{L}_{\text{auto}}^{\text{seq}} = N_{\text{win}} \cdot |\mathcal{P}| \cdot \mathcal{L}_{\text{prop}}$$
RT-NeuS restricts grounding to a fraction $\alpha\rho $of windows (the retained keyframes $\mathcal{K}| = \alpha\rho N_{\text{win}}$
) and performs one batched VLM call per retained window. It also incurs CLIP encoding cost over the full video:
$$\mathcal{L}_{\text{auto}}^{\text{RT-NeuS}} = T \cdot \mathcal{L}_{\text{CLIP}} + \alpha\rho N_{\text{win}} \cdot \mathcal{L}_{\text{VLM}}.$$
Taking the ratio:
$$\frac{\mathcal{L}_{\text{auto}}^{\text{seq}}}{\mathcal{L}_{\text{auto}}^{\text{RT-NeuS}}} = \frac{N_{\text{win}} \cdot |\mathcal{P}| \cdot \mathcal{L}_{\text{prop}}}{T \cdot \mathcal{L}_{\text{CLIP}} + \alpha\rho N_{\text{win}} \cdot \mathcal{L}_{\text{VLM}}}.$$
Factoring out $\alpha\rho N_{\text{win}} \cdot \mathcal{L}_{\text{VLM}}$ from the denominator and assuming $\mathcal{L}_{\text{prop}} \approx \mathcal{L}_{\text{VLM}}$
$$= \frac{|\mathcal{P}|}{\alpha\rho} \cdot \frac{1}{1 + \frac{T \cdot \mathcal{L}_{\text{CLIP}}}{\alpha\rho N_{\text{win}} \cdot \mathcal{L}_{\text{VLM}}}}.$$
The leading factor $\frac{|\mathcal{P}|}{(\alpha\rho)}$ captures the joint contribution of batching ($|\mathcal{P}|$) and adaptive sampling ($\alpha\rho$). The correction factor accounts for CLIP encoding overhead, which is small when CLIP per-frame cost is much less than $\alpha\rho \cdot \mathcal{L}_{\text{VLM}}$ scaled by the window count.

\section{Empirical Retention Rates and Latency Bound Verification}
\label{appendix:retention}
\begin{table}[t]
\centering
\small
\setlength{\tabcolsep}{5pt}
\begin{tabular}{lcccccc}
\toprule
\textbf{Video Length} & $\alpha$ & $\rho$ & $\alpha\rho$ & \textbf{Predicted} & \textbf{Observed} \\
\midrule
15 (Short)         & 0.957 & 0.425 & 0.407 & $\sim 9.8\times$  & $1.77\times$  \\
60 (Medium)        & 0.664 & 0.473 & 0.275 & $\sim 14.5\times$ & $3.16\times$  \\
600 (Medium-Long)  & 0.462 & 0.561 & 0.255 & $\sim 15.7\times$ & $10.52\times$ \\
3600 (Long)        & 0.263 & 0.734 & 0.181 & $\sim 22.1\times$ & $14.58\times$ \\
\midrule
\textbf{Global Average} & \textbf{0.449} & \textbf{0.681} & \textbf{0.194} & \textbf{$\sim 20.6\times$} & \textbf{13.19$\times$} \\
\bottomrule
\end{tabular}
\caption{Empirical retention rates and speedup comparison. 
Predicted speedup is $\frac{|\mathcal{P}|}{(\alpha\rho)}$ with mean $|\mathcal{P}| = 4$. Observed speedup is end-to-end from Table~\ref{tab:latency}. The predicted-observed gap narrows for longer videos, where automaton construction dominates total latency; on shorter videos, $\mathcal{L}_{\text{fixed}}$ 
widens the gap.}
\label{tab:retention-speedup}
\end{table}
\noindent\textbf{Empirical retention rates and speedup verification.} Table~\ref{tab:retention-speedup} reports per-stage retention rates on LongVideoBench. Semantic retention $\alpha$ drops from 0.957 (15s) to 0.263 (3600s) as longer videos contain more irrelevant content. Keyframe retention $\rho$ rises from 0.425 to 0.734 over the same range, since frames surviving Stage~1 in longer videos are already visually distinct. Joint retention $\alpha\rho$ decreases monotonically from 0.407 (15s) to 0.181 (3600s), driving the length-dependent speedup in Table~\ref{tab:latency}. Substituting global mean $\alpha\rho = 0.194$ and $|\mathcal{P}| \approx 4$ into Proposition~\ref{prop:speedup} yields a predicted grounding-stage speedup of $\sim 20.6\times$, reducing to $\sim 14.6\times$ after CLIP-overhead correction. The observed $13.19\times$ falls slightly below, with the residual gap due to $\mathcal{L}_{\text{fixed}}$ (LQ2TL, model checking, final VQA) that the cascade does not reduce.
\end{document}